\pdfoutput=1
\documentclass[journal]{IEEEtran}
\usepackage{amsmath,amsfonts}
\usepackage{algorithmic}
\usepackage[ruled,vlined]{algorithm2e}
\usepackage{array}
\usepackage[caption=false,font=normalsize,labelfont=sf,textfont=sf]{subfig}
\usepackage{textcomp}
\usepackage{stfloats}
\usepackage{color}
\usepackage{url}
\usepackage{verbatim}
\usepackage{graphicx}
\usepackage{cite}
\usepackage{hyperref}
\begin{document}
\renewcommand\topfraction{.9}
\setcounter{totalnumber}{50}
\setcounter{topnumber}{50}
\title{Multi-Task Learning-Enabled Automatic Vessel Draft Reading for Intelligent Maritime Surveillance}
\author{Jingxiang Qu, Ryan Wen Liu, \IEEEmembership{Member, IEEE}, Chenjie Zhao, Yu Guo, Sendren Sheng-Dong Xu,
	
Fenghua Zhu, \textit{Senior Member, IEEE}, and Yisheng Lv, \textit{Senior Member, IEEE}
\thanks{Jingxiang Qu, Ryan Wen Liu, Chenjie Zhao, and Yu Guo are with the Hubei Key Laboratory of Inland Shipping Technology, School of Navigation, Wuhan University of Technology, Wuhan 430063, China, and also with the State Key Laboratory of Maritime Technology and Safety, Wuhan, 430063, China  (e-mail: \{qujx, wenliu, cjzhao, yuguo\}@whut.edu.cn).}%
\thanks{Sendren Sheng-Dong Xu is with the Automation and Control Center and the Graduate Institute of Automation and Control, National Taiwan University of Science and Technology, Taipei 106335, Taiwan (e-mail: sdxu@mail.ntust.edu.tw)}%
\thanks{Fenghua Zhu and Yisheng Lv are with the State Key Laboratory for Management and Control of Complex Systems, Institute of Automation, Chinese Academy of Sciences, Beijing 100190, China (e-mail: \{fenghua.zhu, yisheng.lv\}@ia.ac.cn).}}

\maketitle

\begin{abstract}
  The accurate and efficient vessel draft reading (VDR) is an important component of intelligent maritime surveillance, which could be exploited to assist in judging whether the vessel is normally loaded or overloaded. The computer vision technique with an excellent price-to-performance ratio has become a popular medium to estimate vessel draft depth. However, the traditional estimation methods easily suffer from several limitations, such as sensitivity to low-quality images, high computational cost, etc. In this work, we propose a multi-task learning-enabled computational method (termed MTL-VDR) for generating highly reliable VDR. In particular, our MTL-VDR mainly consists of four components, i.e., draft mark detection, draft scale recognition, vessel/water segmentation, and final draft depth estimation. We first construct a benchmark dataset related to draft mark detection and employ a powerful and efficient convolutional neural network to accurately perform the detection task. The multi-task learning method is then proposed for simultaneous draft scale recognition and vessel/water segmentation. To obtain more robust VDR under complex conditions (e.g., damaged and stained scales, etc.), the accurate draft scales are generated by an automatic correction method, which is presented based on the spatial distribution rules of draft scales. Finally, an adaptive computational method is exploited to yield an accurate and robust draft depth. Extensive experiments have been implemented on the realistic dataset to compare our MTL-VDR with state-of-the-art methods. The results have demonstrated its superior performance in terms of accuracy, robustness, and efficiency. The computational speed exceeds 40 FPS, which satisfies the requirements of real-time maritime surveillance to guarantee vessel traffic safety.
\end{abstract}


\section{Introduction}
\label{sec:introduction}
  \IEEEPARstart{T}{he} rapid growth of artificial intelligence (AI) and the Internet of Things (IoT) has led to significant advancements in intelligent vision technology, which has substantially improved the efficiency and safety of Intelligent Transportation System (ITS). In waterborne transportation systems, visual sensors have been deployed on various maritime surveillance platforms, including shore-based monitoring systems \cite{7536603}, buoys \cite{5597406}, and unmanned surface vessels \cite{9605202}, for safety management purposes. In addition, numerous maritime surveillance tasks can be automated by processing captured visual data, e.g., vessel detection \cite{8911242}, vessel tracking \cite{6165365}, and vessel license plate recognition \cite{9857615}. These automated tasks show the potential benefits of intelligent vision technology in augmenting efficiency and safety in maritime surveillance.
  
  %

\begin{figure}[t]
    \centering
    \setlength{\abovecaptionskip}{0.1cm}
    \includegraphics[width=1.0\linewidth]{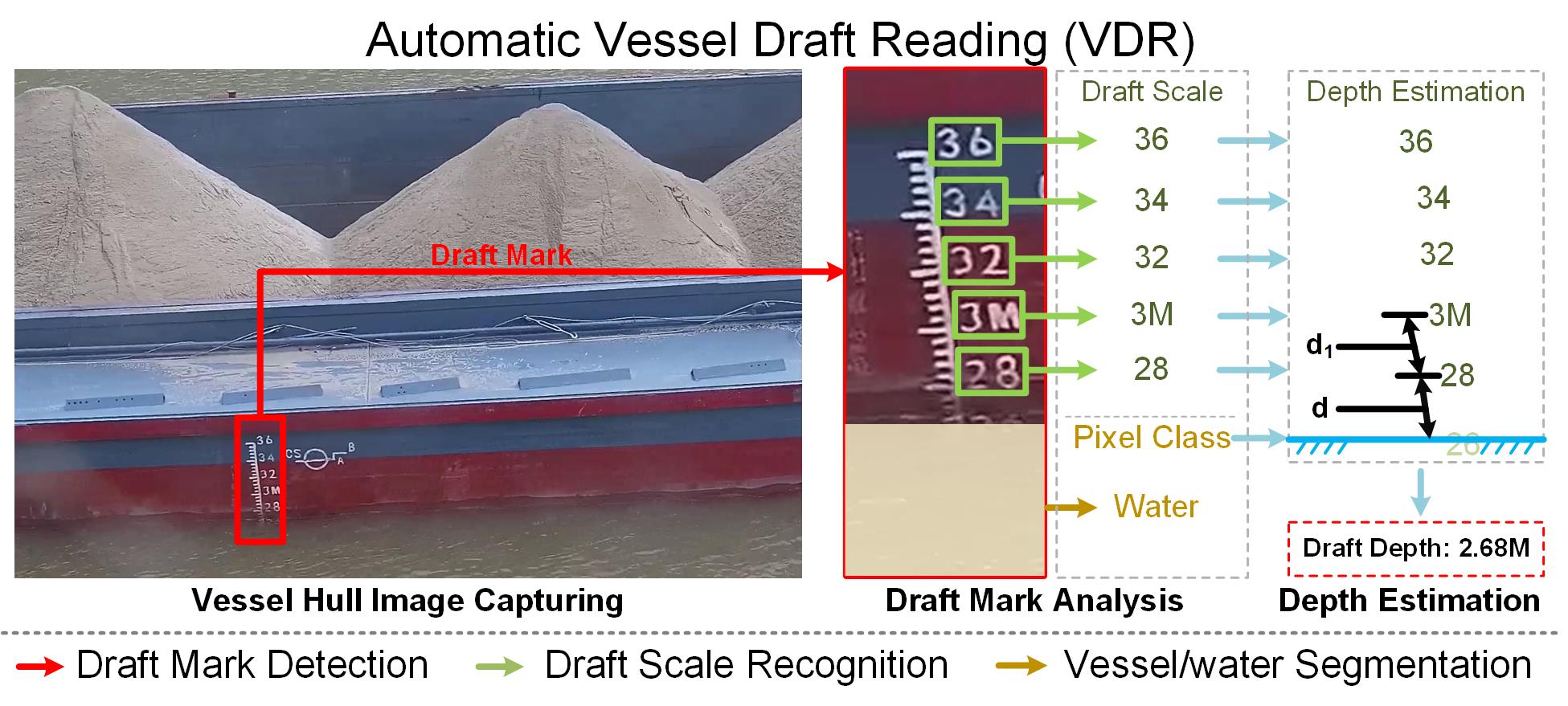}
    \caption{The purpose of VDR is to estimate the draft depth through the captured image. It consists of several subtasks, i.e., draft mark detection, draft scale recognition, vessel/water segmentation, and final draft depth estimation. }
    \label{Figure_draft_reading}
  \end{figure}

  The vessel draft depth is a crucial reference index that estimates the loading capacity of vessels. It serves as the basis for processing claims, port usage fees, and customs clearance taxes \cite{liu2022column}. Moreover, in inland rivers, the complex hydrological environment increases the risk of vessel grounding or sinking, which poses a danger to personnel, damages the vessel, and obstructs waterborne traffic. Vessel draft depth also plays a critical role in preventing these issues. In practical maritime surveillance, management departments deploy transportation boats to inspect the vessel draft mark, leading to significant resource and manpower waste \cite{li2022research}. Some methods have been proposed for automatic VDR. However, the computational speed is slow, which cannot satisfy the requirements of real-time maritime surveillance \footnote{In practical applications, the frame rate of the maritime surveillance camera is around 20 to 30 FPS. Therefore, 30 FPS is regarded as the basic requirement of real-time maritime surveillance.}. To address this issue, we propose a multi-task learning-enabled automatic vessel draft reading method (termed MTL-VDR). It divides the reading process into several subtasks, i.e., draft mark detection, draft scale recognition, vessel/water segmentation, and final draft depth estimation, as illustrated in Fig. \ref{Figure_draft_reading}. 
  
  
  %
  %

  \textbf{Draft Mark Detection.} The draft mark is the pattern to indicate the real-time vessel draught, which is typically depicted at six positions on a vessel, including the bow, midship, and stern on both sides. The initial step of vessel draft reading (VDR) is detecting these draft marks from the captured images of vessel hulls. Existing object detection technologies can accurately solve this problem. In this paper, we employ YOLOv8 \cite{Jocher_YOLO_by_Ultralytics_2023}, the state-of-the-art network for object detection tasks in the computer vision domain. It retains the advanced design in YOLOX \cite{ge2021YOLOx} and YOLOv7 \cite{wang2022YOLOv7}, which enables it to detect draft marks with different sizes. After training on the constructed VDR dataset, it can accurately extract the image patches of draft marks from a 1920 $\times$ 1080 image within 0.03 seconds. 
  %
  
  \textbf{Draft Scale Recognition.} 
  Draft scales are the characters representing the quantitative draught of a vessel, and their recognition is a significant challenge in realizing the VDR. Optical character recognition (OCR) technology has been extensively studied in the field of computer vision. However, recognizing the characters of draft scales presents challenges such as tilt, stains, and low resolution due to various capturing angles, zooming factors, and navigation environments. Current OCR methods thus perform unsatisfactorily. To address this issue, we construct a benchmark dataset and employ the object detection method for draft scale recognition. Our approach significantly improves recognition accuracy compared with existing OCR methods. Moreover, maritime management departments strictly regulate the depiction of vessel draft marks. To further improve the performance of VDR, we propose a draft scale correction method based on their spatial distribution rules. It offers more accurate draft scale recognition results for the final draft depth estimation.
  %
  
  \textbf{Vessel/Water Segmentation.} 
  Vessel/water segmentation is an important task to categorize pixels of the vessel and water to detect the waterline, which represents the boundary between them. Some methods utilize the edge features, employing the Canny operator \cite{canny1986computational} and Hough transformation for waterline detection \cite{ran2012draft, tsujii2016automatic}. However, these methods are unfriendly to curves and sensitive to color changes on the hull, which are unstable in different environments. Deep learning has enabled significant improvements in image segmentation tasks. However, most segmentation methods require powerful computational devices due to fine pixel-level calculations, and their color sensitivity may cause them to mistake vessel hull stains for water. In this paper, we propose a multi-task learning network that achieves draft scale recognition and vessel/water segmentation simultaneously. It greatly improves the efficiency while maintaining competitive performance on both subtasks. Additionally, we design a specialized loss function to improve the accuracy of waterline detection, which provides more accurate locations of waterlines for the final draft depth estimation.
  %

  After completion of the above subtasks, we propose a dynamically adaptive method that uses relative distance or character height to estimate the draft depth based on the recognition conditions of the draft scale. In general, our MTL-VDR utilizes multi-task learning, a novel loss function, and prior knowledge to address the problems of complex computation, unsatisfactory accuracy, and poor robustness to stains that are common in current automatic VDR methods. The contributions of this paper are summarized as follows.
  %
    \begin{itemize}
    \item A multi-task learning network for draft scale recognition and vessel/water segmentation is proposed. It significantly improves the reading efficiency with satisfactory accuracy. Besides, a loss function focusing on the waterline location is suggested to improve the segmentation performance on pixels of waterlines.
    \item A draft scale correction method based on prior knowledge about the spatial distribution rules is designed. It successfully improves the robustness of the proposed MTL-VDR to stains on vessel hulls.
    \item We construct a benchmark dataset for multiple subtasks of VDR, including draft mark detection, draft scale recognition, vessel/water segmentation, and draft depth estimation. 
    \end{itemize}

    The main benefit of the proposed MTL-VDR is that it utilizes a multi-task learning network with a newly designed loss function, which improves the efficiency and accuracy of VDR. Subsequently, the issue of hull stain is addressed through draft scale correction and dynamically adaptive draft depth estimation methods. Additionally, the efficiency, effectiveness, and robustness of the proposed method have been verified on a newly-developed VDR dataset, demonstrating the practical applicability of our MTL-VDR.
    %
    %
    
    The rest of this paper is organized as follows: Section \ref{related_work} briefly reviews previous relevant technologies about draft reading tasks. The detailed implementations of the proposed method are presented in Section \ref{proposed_method}, including draft mark detection, multi-task learning network, draft scale recognition and correction, and draft depth estimation. A new-developed dataset for the vessel draft reading task and extensive experiments are introduced in Section \ref{experiment}. Finally, we conclude our work and discuss future perspectives in Section \ref{conclusion}.

\section{Related Work}
\label{related_work}
  %
  %
  This section mainly introduces recent studies related to object detection, multi-task learning, and vessel draft reading.
  \subsection{Object Detection}
   The suggested draft reading approach extracts draft marks and recognizes draft scales via object detection technology, which is widely researched in the computer vision domain. With the development of parallel computation devices like the graphics processing unit (GPU) and tensor processing unit (TPU), deep learning-based object detection methods have shown great competitiveness, which can be broadly classified into two-stage and one-stage methods. Two-stage methods, e.g., Region-CNN \cite{girshick2014rich} and its variants \cite{girshick2015fast, ren2015faster}, regard the detection as a “coarse-to-fine” process \cite{zou2019object}, using separate steps to accomplish the detection. One-stage methods try to complete the detection tasks with a single network. You only look once (YOLO) was proposed in 2015, which is the first deep learning-based one-stage method \cite{redmon2016you}. It divides the image into regions and predicts bounding boxes with probability for each region simultaneously. In the subsequent years, YOLO has been researched and updated into multiple versions, including YOLOv3 \cite{redmon2018YOLOv3}, YOLOv4 \cite{bochkovskiy2020YOLOv4}, YOLOX \cite{ge2021YOLOx}, etc. The state-of-the-art version is YOLOv8\cite{Jocher_YOLO_by_Ultralytics_2023}, which is employed for draft mark detection in our approach. Compared with two-stage methods, one-stage methods sacrifice some detection accuracy for better efficiency, making them more suitable for real-time tasks. 
   Detection anchors are pre-defined bounding boxes that are placed regularly on feature maps and determine the size of output bounding boxes. However, the visual size of draft marks can be diverse due to different imaging conditions, making it difficult for anchor-based object detection methods (e.g., SSD \cite{liu2016ssd}, RetinaNet \cite{lin2017focal}, and YOLOv7 \cite{wang2022YOLOv7}) to generate universal anchors. The anchor-free methods (such as CornerNet \cite{law2018cornernet}, FCOS \cite{tian2019fcos}, and YOLOX \cite{ge2021YOLOx}) are more suitable for VDR tasks because they do not rely on pre-defined anchors and can handle diverse visual sizes. Therefore, our MTL-VDR utilizes one-stage anchor-free methods to achieve draft mark detection and draft scale recognition. It provides an adequate and efficient solution that satisfies the requirements of real-time maritime surveillance.
    \begin{figure}[t]
    	\centering
    	\setlength{\abovecaptionskip}{0.1cm}
            \vspace{0cm}
    	\includegraphics[width=1.0\linewidth]{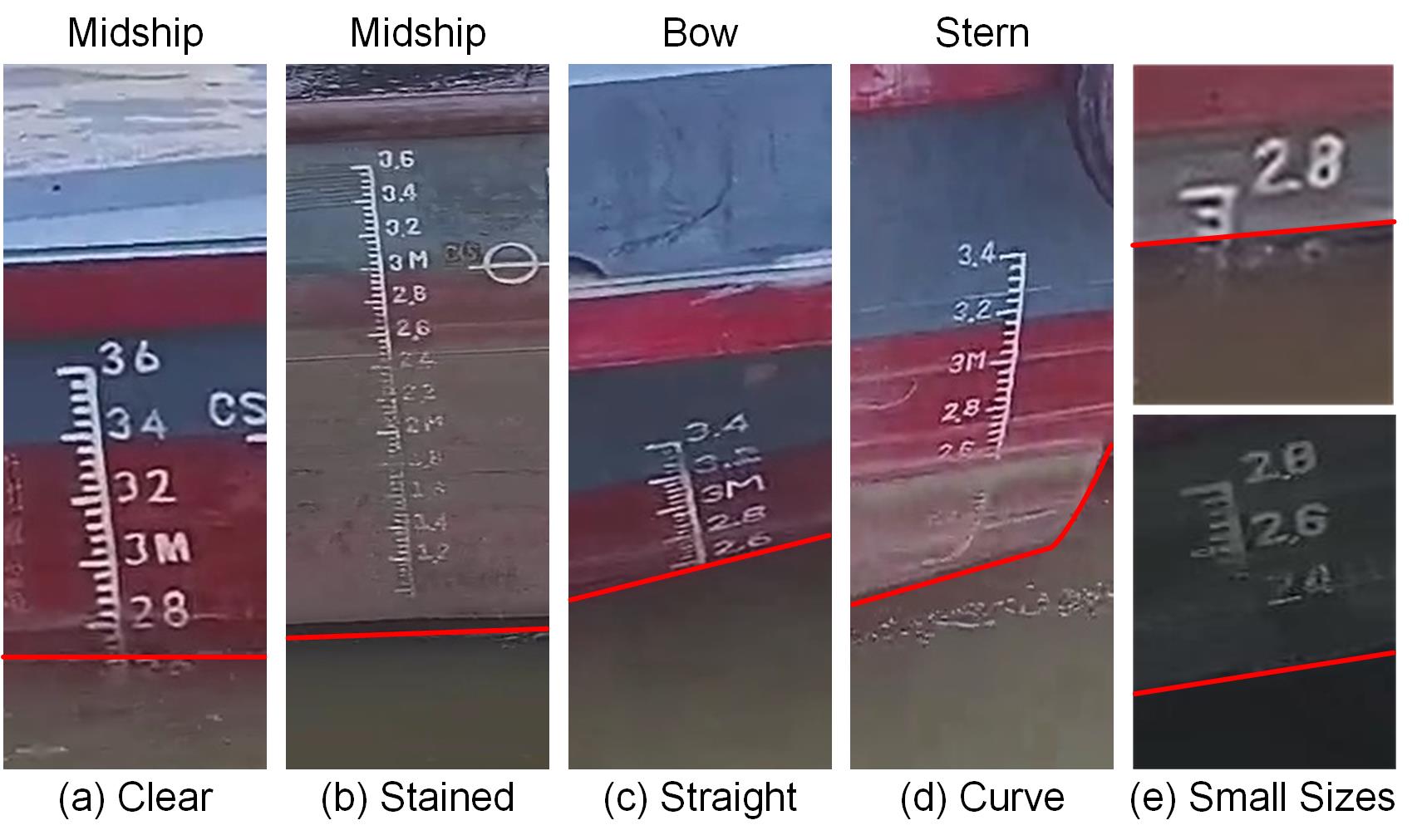}
    	\caption{The challenges in practical VDR tasks, including irregular shapes of waterlines (the red lines depicted in the figures), vessel hull stains, small visual sizes, etc.}
    	\label{Figure_draft_types}
    \end{figure}
    \begin{figure}[t]
    	\centering
    	\setlength{\abovecaptionskip}{0.1cm}
            \vspace{-0cm}
    	\includegraphics[width=1.0\linewidth]{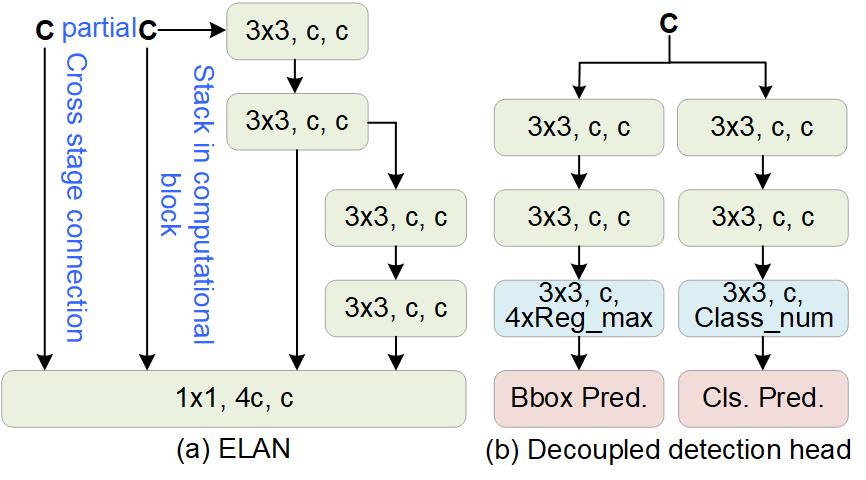}
    	\caption{The diagrams of ELAN \cite{wang2022YOLOv7} structure and decoupled detection heads in YOLOv8 \cite{Jocher_YOLO_by_Ultralytics_2023}. ELAN enables the backbone to extract more gradient features, and the decoupled detection head provides more accurate detection results compared with the normal head \cite{ge2021YOLOx}.}
    	\label{Figure_YOLOv8}
    \end{figure}
  \subsection{Multi-Task Learning}
    The draft scale recognition and vessel/water segmentation tasks are implemented on a single image. In our method, they are completed by a multi-task learning network to reduce the computational complexity. After the rebirth of neural networks, deep learning has been widely used in a variety of visual tasks, such as image dehazing \cite{9764372}, object detection \cite{9444810}, and instance segmentation \cite{9556571}. The majority of them employ a single network to address a particular issue. However, when multiple visual tasks need to be accomplished on a single image, separate networks will undoubtedly cause complex calculations with poor efficiency. Multi-task learning technology has been thus widely used in the industrial field. For instance, Wang \textit{et al}. \cite{wang2022explainable} tried to employ the multi-task learning network in medical diagnostics, and Liu \textit{et al}. \cite{liu2021multi} designed a multi-task CNN for vessel detection, classification, and segmentation. It successfully improves the efficiency and accuracy of maritime object detection. Moreover, numerous multi-task learning methods are proposed for improving performance by sharing complementary information or acting as a regularizer for one another task \cite{9336293}. For instance, Zhou \textit{et al}. \cite{zhou2022enhanced} improved the accuracy of detecting pedestrians at far distances through multi-task learning about both semantic segmentation and object detection. Wu \textit{et al}. \cite{wu2020multi} adopted the multi-task learning mechanism to study the multi-timescale forecast. In our method, we employ multi-task learning technology for different subtasks, i.e., draft scale recognition and vessel/water segmentation. It has strict requirements for the balance between the loss functions of different tasks. During the training period, we thus employ a dynamically weighted loss function based on the homoscedastic uncertainty of each task. It adjusts the weights of the loss functions accordingly by measuring their variance at different training stages.
  \subsection{Vessel Draft Reading}
    The manual measurement of vessel draft is a knotty and laborious task. With the advancement of intelligent visual technology, several visual data-based automatic VDR methods have been proposed. Some of these methods estimate the waterline based on edge features by employing techniques such as Canny operator-based edge detection and Hough transformation. For instance, Ran \textit{et al}. \cite{ran2012draft} used these techniques to estimate the waterline. Similarly, Tsujii \textit{et al}. \cite{tsujii2016automatic} detected the draft mark using morphology from gray-scale images and then used the Canny operator to identify the possible boundaries of the waterline. However, due to the different camera angles in many cases, the waterline is usually an irregular curve, making Hough transformation unsuitable for waterline detection, as shown in Fig. \ref{Figure_draft_types} (d). Additionally, when hulls are stained or undergo color changes (as depicted in Fig. \ref{Figure_draft_types} (b)), additional gradient features will affect the outcome of Canny operator-based edge detection. To address these issues, Liu \textit{et al}. \cite{liu2022column} proposed a novel column-based waterline detection method, which achieves VDR with an error of less than 0.3 meters. Nevertheless, this method is not robust enough in scenarios where the vessel hull stains have a similar color to the water. Furthermore, their approach only recognizes integral scales, leading to significant deviations in draft depth estimation. Li \textit{et al}. \cite{li2022research} suggested using separate networks to complete different subtasks of VDR, which improved accuracy but resulted in complex computations from multiple deep learning models. 

    Although some studies have been proposed for automatic vessel draft reading, it is still challenging to achieve a satisfactory balance between accuracy and efficiency. Furthermore, the problem of vessel hull stain, which is common in inland vessels, has received little attention in previous works. As such, current methods are not robust enough to handle the complex environments in real-time maritime surveillance. It further highlights the need for accurate, efficient, and robust VDR methods.
\section{MTL-VDR: Multi-Task Learning-Enabled Automatic Vessel Draft Reading}
\label{proposed_method}
  The implementation of the proposed MTL-VDR involves several steps. Firstly, image patches containing draft marks are extracted from raw visual data. Subsequently, a multi-task learning network is developed to perform draft scale recognition and vessel/water segmentation in a single image. The recognized scales are then fed into the spatial distribution rule-based correction method to ensure their recognition and positional accuracy. Finally, we provide a dynamically adaptive method for estimating the final draft depth.
    \begin{figure*}[t]
    	\centering
    	\setlength{\abovecaptionskip}{0.1cm}
    	\includegraphics[width=1.0\linewidth]{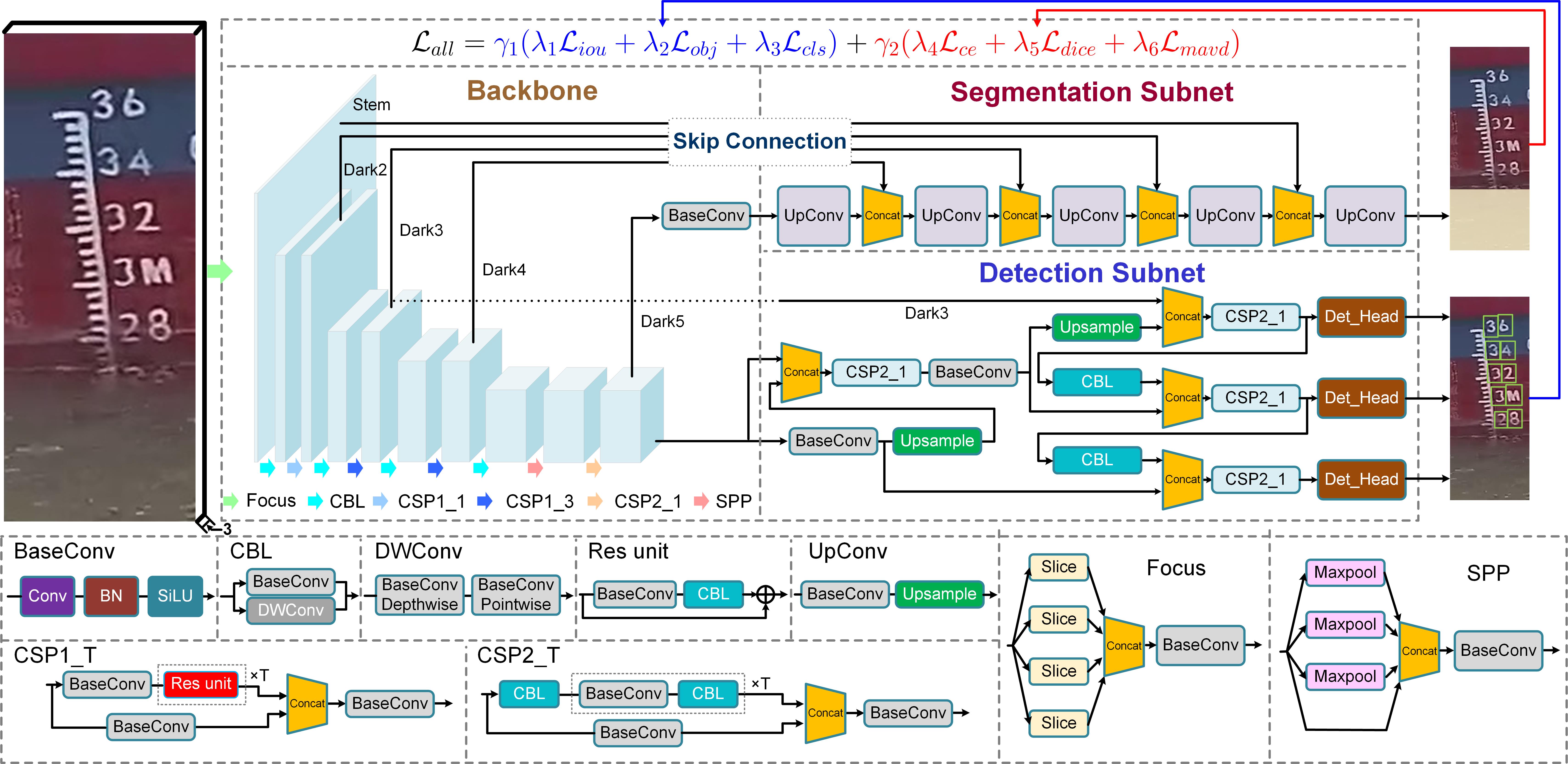}
    	\caption{The architecture of the proposed multi-task learning network. After the backbone, the multi-scale feature maps are separately fed into two subnets for draft scale character detection and vessel/water segmentation. A joint loss function is then designed to simultaneously supervise two tasks, whose weights are dynamically adjusted during the training period.}
    	\label{Figure_Architecture}
    \end{figure*}
  \subsection{Draft Mark Detection}
    The first step of automatic VDR is extracting image patches with draft marks from raw visual data. To facilitate this process, we constructed a draft mark detection dataset comprising 2226 images with different categories of vessels, e.g., cargo vessels, passenger vessels, patrol vessels, etc. As shown in Fig. \ref{Figure_draft_types}, the size of draft marks can vary significantly across different vessel categories, posing a challenge for accurate detection. To address this issue, we employed YOLOv8 \cite{Jocher_YOLO_by_Ultralytics_2023} as the baseline for the draft mark detection network and trained it on the proposed dataset. YOLOv8 is the latest version of YOLO and has demonstrated satisfactory performance on public datasets. It achieves accurate and efficient object detection with the retention of the ELAN structure from YOLOv7 \cite{wang2022YOLOv7} and the decoupled detection head from YOLOX \cite{ge2021YOLOx}, as illustrated in Fig. \ref{Figure_YOLOv8}, where $c$ represents the channels of the feature map or convolutional kernel. Moreover, it performs satisfactorily on the draft marks with different sizes due to the anchor-free strategy. On the testing data, it achieves over 85\% accuracy, which is sufficient for industrial applications. After detection, the image patch of the draft mark is fed into the proposed multi-task learning network for draft scale recognition and vessel/water segmentation.
  \subsection{Multi-Task Learning Network}
    The architecture of the proposed multi-task learning network is depicted in Fig. \ref{Figure_Architecture}. We use DarkNet-53 \cite{redmon2018YOLOv3}, a multi-scale convolution neural network as the backbone to extract the feature maps. The basic unit of the backbone is BaseConv, consisting of the convolutional layer $C$, batch-normalization layer $B_n$, and SiLU function $A_f$. The output $F^o$ of the BaseConv can be given by
    \begin{equation}
        F^o = A_{f}(B_{n}(C(\mathcal{I}))),
    \end{equation}
    where $\mathcal{I}$ is the input of the BaseConv unit. The original images are first split and concatenated by the Focus unit. The feature maps are then downsampled via the convolutional layer in BaseConv units, whose kernel size and stride are set to 3 and 2, respectively. The extracted features are stored in the feature list $L_F$, i.e.,
    \begin{equation}
        L_F = [D_1, D_2, D_3, D_4, D_5],
    \end{equation}
    where $D_{1 \leq i \leq 5}$ are the multi-scale feature maps, respectively. They are then connected to the detection and segmentation subnets.

    After the backbone, we design two subnets for different tasks. The size of captured draft scale characters is diversified due to different imaging conditions. Therefore, the anchor-based object detection method works unsatisfactorily on the recognition task. In our MTL-VDR, the baseline of the detection subnet is YOLOX-s \cite{ge2021YOLOx}, an anchor-free object detection method including multiple detection heads. Specifically, each detection head consists of multiple branches for different tasks, i.e., classification, regression, and intersection over union (IoU) prediction. The IoU branch predicts whether each location $(x, y)$ in the feature map is the background or object. The regression branch generates a four-dimensional vector containing information about the bounding box. The classification branch calculates the probability of the classes, including the numbers '0'-'9' and the unit 'M'.

    Compared with other segmentation tasks, vessel/water segmentation is relatively simple, which only needs to distinguish the water from other pixels. Therefore, in the constructed dataset, the pixels of water are annotated as the object, and others are the background. In the segmentation subnet, inspired by \cite{ronneberger2015u}, we design a u-net structure consisting of the UpConv units with the skip connection. Specifically, the multi-scale feature maps are generated by the shared backbone, DarkNet-53, which extracts more information compared with the original u-net. Each feature map is then fed to UpConv units for skip connection and upsampling, which can be given by
    \begin{equation}
        F^u = (U(F^o) ; D_n),
    \end{equation}
    where $F^u$ represents the upsampled feature map, $U (\cdot)$ is the upsample layer, $(;)$ represents the concatenation operator, and $D_n$ denotes the corresponding feature map connected from the backbone.

    To comprehensively supervise the learning period, we design a joint loss function $\mathcal{L}_{all}$, which can be given by
    \begin{equation}
        \mathcal{L}_{all} = \gamma_1 \mathcal{L}_{det} + \gamma_2 \mathcal{L}_{seg},
    \end{equation}
    where $\mathcal{L}_{det}$ and $\mathcal{L}_{seg}$ represent the loss of detection and segmentation tasks, respectively. Inspired by \cite{kendall2018multi}, the weight of each loss, i.e., $\gamma_1$ and $\gamma_2$, is dynamically adjusted during the training period considering the homoscedastic uncertainty of each task. It achieves a satisfactory balance between the learning abilities of different tasks. The $\mathcal{L}_{det}$ consists of the loss function about the classification, regression, and IoU, which can be given by
    \begin{equation}
    \label{L_det}
        \mathcal{L}_{det} = \lambda_1 \mathcal{L}_{iou} + \lambda_2 \mathcal{L}_{obj} + \lambda_3 \mathcal{L}_{cls},
    \end{equation}
    where $\lambda_{1 \leq i \leq 3}$ are the weights set to the same as YOLOX \cite{ge2021YOLOx}, i.e., 5, 1, and 1, respectively. In Eq. (\ref{L_det}), $\mathcal{L}_{obj}$ and $\mathcal{L}_{cls}$ are the corresponding binary cross entropy (BCE) loss for regression and classification tasks, respectively. $\mathcal{L}_{iou}$ will calculate the IoU between the annotated and predicted bounding boxes, which can be expressed as
    \begin{equation}
        IoU (P, L) = \frac{P \cap L}{P \cup L},
    \end{equation}
    where $P$ and $L$ are the bounding boxes of the prediction and ground truth, respectively. 
    
    For the vessel/water segmentation task, we first employ the cross entropy loss ($\mathcal{L}_{ce}$) and dice loss ($\mathcal{L}_{dice}$) \cite{milletari2016v}  to supervise the network. Moreover, the top water pixels, which determine the location of the waterline, are exceedingly important for draft depth estimation. Therefore, we calculate the mean absolute vertical distance (MAVD) between the prediction and label as a special loss function $\mathcal{L}_{mavd}$, i.e.,
    \begin{equation}
      \label{L_MAVD}
      \mathcal{L}_{mavd} =  \frac{1}{W}\sum_{0<w<W} \left| y^{label}_{w} - y^{pre}_{w} \right|,
     \end{equation}
    where $W$ is the width of the draft image. $y^{label}_{w}$ and $y^{pre}_{w}$ represent the vertical coordinates of top water pixels in the column $w$ of label and prediction, respectively. Compared with classical loss functions for image segmentation, $\mathcal{L}_{mavd}$ pays more attention to the location of the waterline. It enables the network to perform better on the pixels of the boundary between the vessel and water. In general, the overall loss function of the segmentation task can be given by
    \begin{equation}
        \mathcal{L}_{seg} = \lambda_4\mathcal{L}_{ce} + \lambda_5\mathcal{L}_{dice} + \lambda_6\mathcal{L}_{mavd},
    \end{equation}
    where $\lambda_{4 \leq i \leq 6}$ are the weights of each loss function which are set to 1, simultaneously. 
    %
  \subsection{Draft Scale Recognition and Correction}
    \begin{figure}[t]
    	\centering
    	\setlength{\abovecaptionskip}{0.1cm}
    	\includegraphics[width=1.0\linewidth]{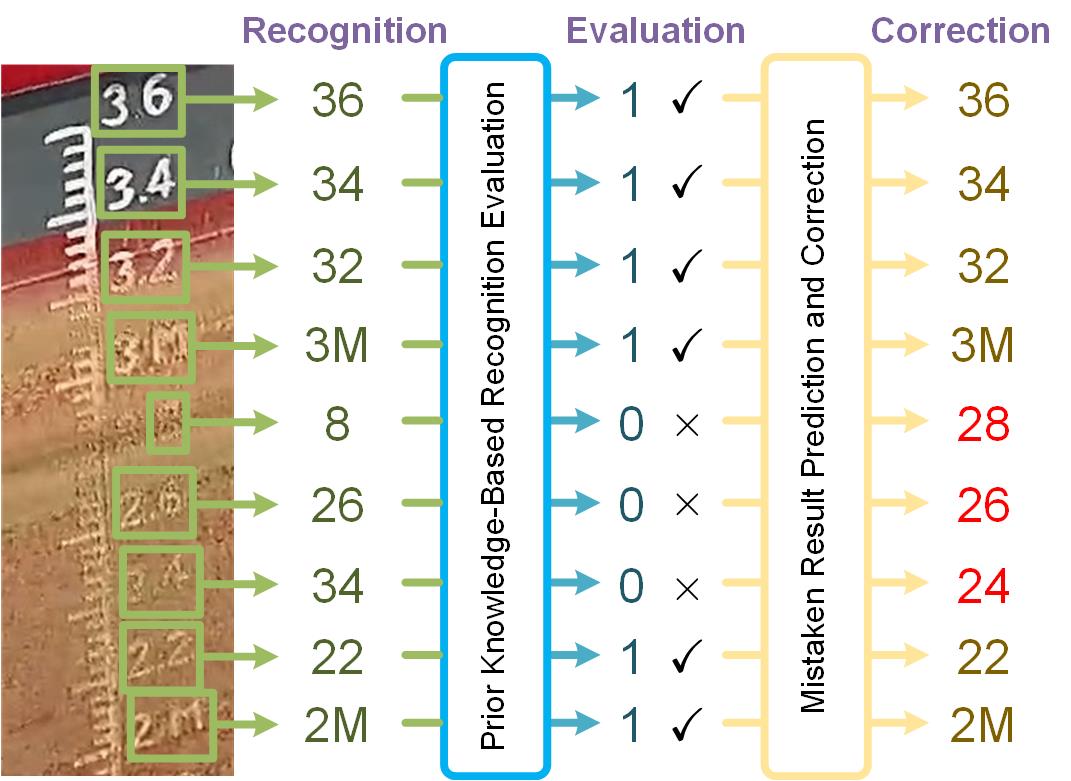}
    	\caption{The draft scale recognition results are initially assessed based on spatial distribution rules. The scales obeying the distribution rules will be scored, while the others are scoreless. For the scoreless scales, we will replace them with the correct recognition results, which are predicted referring to the scored scales. }
    	\label{Figure_Recognition_Correction}
    \end{figure}
    \begin{algorithm}[!htb]
		\footnotesize
		\caption{Draft Scale Recognition and Correction}
		\label{al:draft_mark_correction}
		\LinesNumbered
		\KwIn {
		    $Bboxes = \{[Bbox_{1}, Class_1, Conf_1],..., [Bbox_{i}, Class_i, Conf_i],..., $ $[Bbox_{n}, Class_n, Conf_n]\}$: The outputs of the detection subnets in the proposed multi-task learning network, which consists of the bounding box $Bbox$,  classification result $Class$, and confidence $Conf$;
			}
		\KwOut {$\mathcal{D}_{correct}: \{[D_{1},..., D_{i},..., D_{n}\} $: All of recognized draft scales with their locations, which are sorted from top (large) to bottom (small)} 
		\textbf{Initialization:} $Bbox = [x, y, w, h]$: The set of center point and shape size of the bounding box\;
		$Class = \{'0'-'9', 'M'\}$: The classes of the draft character detection results, including the number '0'-'9' and the unit 'M'\;
		$D = [x_{c}, y_{c}, h, c, s]$: The set of recognized draft scale, consisting of the horizontal coordinate $x_{c}$, vertical coordinate $y_{c}$, character pixel height $h$, scale characters $c$, and the assessed score $s$\;
		$IoU (Bbox_i, Bbox_j)$: The function to calculate the IoU between $Bbox_i$ and $Bbox_j$\;
		$\mathcal{C}( ; )$ : The function of string concatenation and replacing the unit 'M' to the number '0'\;
		$\mathcal{F}(\mathcal{D}_{correct}, D_i) = [D_L, D_N]$ : The function for searching two nearest correct recognition results ($D_L$ and $D_N$) of the $D_i$ in the $\mathcal{D}_{correct}$\;
		$\phi (x)$: The function to find the closest integral multiple of 0.2 to $x$, which should not be duplicate with the existing scales.
		\\
		\tcp{Step 1. Cross-class non-maximum suppression.}
		\For{$[Bbox_i, Class_i, Conf_i]$ in $Bboxes$}{
		  \For{$[Bbox_j, Class_j, Conf_j]$ in $Bboxes$}{
		    \uIf{$IoU(Bbox_i, Bbox_j) > 0.3$ and $Conf_i > Conf_j$}{
		      Delete $[Bbox_j, Class_j, Conf_j]$ from $Bboxes$\;
		      }}
		  }
		\tcp{Step 2. Draft scale recognition (generating the scales through associating the detection results).}
		\For{$[Bbox_i, Class_i, Conf_i]$ in $Bboxes$}{
			\For{$[Bbox_j, Class_j, Conf_j]$ in $Bboxes$}{
				\uIf{$0 < x_j - x_i < 2w_i  $ and $\left| y_j - y_i \right| < \min{(h_i, h_j)}$}{
					$x_{c} = \frac{x_i + x_j}{2}$\;
					$y_{c} = \frac{y_i + y_j}{2}$\;
					$c = \frac{\mathcal{C}(Class_i; Class_j)}{10}$\;
                        $h = \frac{h_i + h_j}{2}$\;
					Add $[x_{c}, y_{c}, h, c, 1]$ to $\mathcal{D}_{correct}$\;}
		}}
		\tcp{Step 3. Recognized results scoring.}
		\For{$[x_{i}, y_{i}, h_i, c_i, s_i]$ in $\mathcal{D}_{correct}$}{
			\uIf{($\left|c_i - c_{i - 1}\right| == 0.2$ and $y_i - y_{i - 1} < 2.3 h_i$) or ($\left|c_i - c_{i + 1}\right| == 0.2$ and $y_{i + 1} - y_i < 2.3 h_i$)}{
			 \textbf{Continue}.
			}
			\Else{
			  $s_i = 0$\;
				}}
		\tcp{Step 4. Mistaken recognition correction.}
		\For{$D_i = [x_{i}, y_{i}, h_i, c_i, s_i]$ in $\mathcal{D}_{correct}$}{
		  \uIf{$s_i == 0$}{
			$[D_L, D_N] = \mathcal{F}(\mathcal{D}_{correct}, D_i)$\;
			$d_1 = y_N - y_L$\;
			$d_2 = y_i - y_L$\;
			$c_i = \phi(c_L - \frac{d_2 \cdot (c_L - c_N)}{d_1})$\;
			$s_i = 1$\\
			}
		  \Else{
			  \textbf{Continue}.
				}
		}
	\end{algorithm}
    The proposed multi-task learning network predicts the locations and classes of draft scale characters. It is necessary to associate them to recognize the specific scales. At first, to prevent repetitive character detection, we implement a cross-class non-maximum suppression based on the spatial distribution rules of the draft scale. Due to the non-overlapping property of the draft scales, for the detection results with more than 30\% IoU, it only persists the detected character with the largest confidence. The duplicate bounding boxes of the same object are thus deleted. Then we generate the draft scales by merging the characters whose vertical and horizontal gaps are less than the character height and width, respectively. However, mistaken recognition often occurs due to the stains or low resolution of the draft image. It always brings serious bias to the final draft depth estimation. To tackle this problem, we design a spatial distribution rule-based draft scale correction method.

    The depiction of inland vessel draft marks must comply with the regulations: a) the order of the draft scales is strictly arranged from large (top) to small (bottom); b) the value difference between adjacent scales is 0.2 meters; c) the height of each draft scale pattern is 0.1 meters. Based on these spatial distribution rules, all recognition results will be judged first, as shown in Fig. \ref{Figure_Recognition_Correction}. The scales will be scored under the following conditions: a) it has a neighbor scale with an absolute value difference of 0.2 meters; b) the position gap between it and this neighbor on the image is less than 2.3 times its character height. The mistaken recognition is scoreless, which will be replaced by the correct recognition results predicted by referring to the scored scales. Specifically, the vertical distance between the two nearest accurate scales ($d_1$), the vertical distance between the nearest accurate scale and the scale being corrected ($d_2$), and the number difference in reading between the scales are all calculated. Then the predicted scales ($c_i$) can be estimated through
    \begin{equation}
      c_i = \phi \left(c_L - \frac{d_2 \cdot (c_L - c_N)}{d_1} \right),
    \end{equation}
    where $\phi (x)$ is the function to find the closest integral multiple of 0.2 to $x$ and avoid the duplication between predicted and existing scales. $c_L$ and $c_N$ are the nearest two correct scales. It is noted that $c_L$ is larger than $c_N$, and $d_2$ should be negative if $c_L$ is smaller than $c_i$. The pseudo code of the proposed draft scale recognition and correction method is presented in Algorithm \ref{al:draft_mark_correction}.
  \subsection{Draft Depth Estimation}
    \begin{figure}[t]
    	\centering
    	\setlength{\abovecaptionskip}{0.1cm}
    	\includegraphics[width=1.0\linewidth]{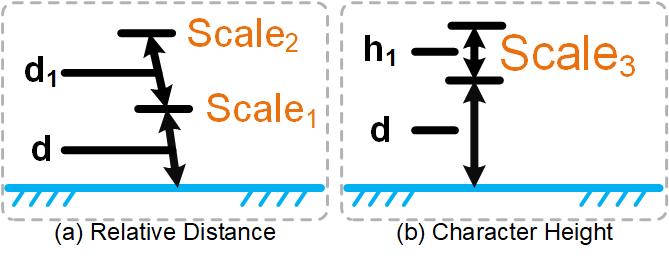}
    	\caption{The diagrams of draft depth estimation methods under different reading conditions. Method (a) can achieve a more accurate estimation when there are many available scales. Method (b) will be utilized when there is only one available scale due to heavy loading or vessel hull stains.}
    	\label{Figure_draft_estimation}
    \end{figure}
    After the draft scale recognition, the next step is to estimate the current draft depth according to the relative positions of the waterline and draft scales. However, due to the stains on vessel hulls and different load capacities, the available scales are limited in many conditions. To improve the universality of our method, we use dynamically adaptive linear fitting methods to estimate the draft depth under different conditions. Specifically, if over two scales are available, the draft depth $\mathcal{D}$ will be estimated via Eq. (\ref{depth_estimation_distance_1}), as depicted in Fig. \ref{Figure_draft_estimation} (a), which can be expressed as
    \begin{equation}
    \label{depth_estimation_distance_1}
        \mathcal{D} = \frac{d}{d_1}(S_1 - S_2),
    \end{equation}
    where $d$ is the relative distance of water to the closest scale ($S_1$ in Fig. \ref{Figure_draft_estimation} (a) and $S_3$ in Fig. \ref{Figure_draft_estimation} (b)), and $d_1$ is the relative distance of $S_1$ to the neighbor scale ($S_2$). When there is only one available scale, the draft depth will be estimated referring to the height of the character, as depicted in Fig. \ref{Figure_draft_estimation} (b), which can be given by
    \begin{equation}
    \label{depth_estimation_distance_2}
        \mathcal{D} = S_3 - \frac{\beta \cdot d}{h_1},
    \end{equation}
    where $h_1$ is the height of the available scale character ($S_3$), and $\beta$ is the realistic height of the scale character, which is 0.1 meters on inland vessels. Compared with the static depth estimation methods in previous works \cite{liu2022column, wang2021computer,tsujii2016automatic}, our method is available for the draft marks under different conditions, e.g., light-loaded vessels, heavy-loaded vessels, clear marks, stained marks, etc.
\section{Experiments and Analysis}
\label{experiment}
  In this section, we briefly introduce the implementation details and evaluation metrics of the experiments. Then we construct a benchmark dataset for VDR. Moreover, we employ some classical and state-of-the-art methods for comparison, which are evaluated from the accuracy, efficiency, and robustness to stains in detail. Finally, we make the ablation study on the necessity of each component in the proposed method.
    \subsection{Implementation Details}
      The proposed multi-task learning network is built on Pytorch and trained for 200 epochs. Specifically, the Adam optimizer is employed for training, and the running time of experimental methods is calculated on a laptop with an AMD Ryzen 7 5800H CPU accelerated by a NVIDIA GTX 3060 GPU. For a fair comparison, the parameters of all competing methods are pre-trained on the PASCAL VOC dataset \cite{everingham2010pascal} and trained for 200 epochs on the constructed dataset. In the experiment, the outputs of the compared methods are fed into the dynamically adaptive linear fitting method for the final draft depth estimation.
    \subsection{Evaluation Metrics}
      \begin{figure}[t]
    	\centering
    	\setlength{\abovecaptionskip}{0.1cm}
    	\includegraphics[width=1.0\linewidth]{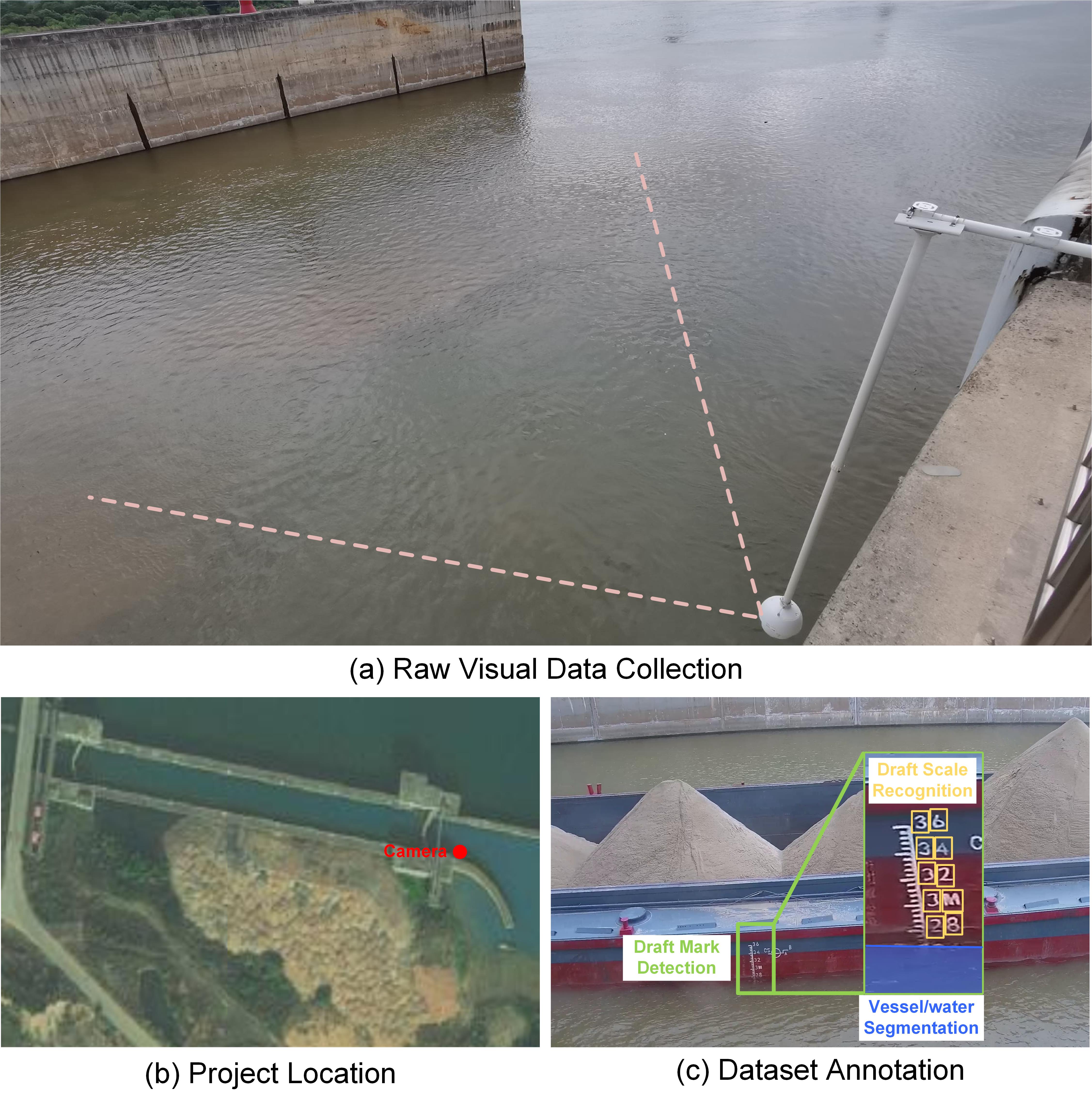}
    	\caption{The camera is extended downward from the side of the lock, observing the vessel hulls horizontally, as shown in Fig. (a). Fig. (b) presents the position where the visual data is taken in Guangdong Province, China. Subsequently, the images are finely annotated for different tasks, as depicted in Fig. (c).}
    	\label{Figure_dataset_collection}
      \end{figure}
      \begin{table}[t]
      	\scriptsize
      	\centering
      	\caption{Details of the constructed VDR dataset serving for different tasks during the VDR processing including draft mark detection, draft scale recognition, vessel/water segmentation, and draft depth estimation. }
      	\label{Tab:Dataset}
            \scriptsize
      	\begin{tabular}{c|c|c|c}
      		\hline
      		Task                                                                & Label Type                                                        & Number & Classes                                                                  \\ \hline \hline
      		\begin{tabular}[c]{@{}c@{}}Draft Mark \\ Detection\end{tabular}      & \begin{tabular}[c]{@{}c@{}}VOC\\ (Detection)\end{tabular}         & 2226   & \begin{tabular}[c]{@{}c@{}}Draft Marks, \\ Vessel Names\end{tabular}     \\ \hline
      		\begin{tabular}[c]{@{}c@{}}Draft Scale \\ Recognition\end{tabular}   & \begin{tabular}[c]{@{}c@{}}VOC\\ (Detection)\end{tabular}         & 1198   & \begin{tabular}[c]{@{}c@{}}Numbers: '0'-'9', \\ Unit: 'M'\end{tabular}   \\ \hline
      		\begin{tabular}[c]{@{}c@{}}Vessel/Water \\ Segmentation\end{tabular} & \begin{tabular}[c]{@{}c@{}}VOC, COCO\\ (Segmentation)\end{tabular} & 1198   & \begin{tabular}[c]{@{}c@{}}Object: 'water', \\ 'background'\end{tabular} \\ \hline
      		\begin{tabular}[c]{@{}c@{}}Draft Depth\\ Esitimation\end{tabular}    & Manual Reading                                                                 & 45     & \begin{tabular}[c]{@{}c@{}}30 Clear, \\ 15 Stained\end{tabular}          \\ \hline
      	\end{tabular}
      \end{table}
      %
      %
      \subsubsection{Mean Absolute Vertical Distance (MAVD)} The location of the waterline greatly impacts the final draft depth estimation. The image segmentation-based waterline detection method regards the top pixels of water as the location of waterlines. Therefore, in this paper, we propose a special metric termed MAVD to evaluate its accuracy. It calculates the absolute distance between the labels and the predicted top water pixels, which can be given by
        \begin{equation}
        \label{MAVD}
        {\rm{MAVD}} =  \frac{1}{W}\sum_{0<w<W} \left| y^{label}_{w} - y^{pre}_{w} \right|.
        \end{equation}

      Theoretically, lower MAVD represents more accurate waterline detection in our experiments.
      \subsubsection{Mean Absolute Draft Depth Error (MADDE)} The final output of the proposed method is the draft depth of the vessel. To quantitatively verify the accuracy, we first manually read and annotate the real depth of some draft marks. The mean error between the prediction and label is calculated through
        \begin{equation}
        \label{MADDE}
        {\rm{MADDE}} = \frac{1}{N}\sum_{0<n<N} \left| \mathcal{D}^{label}_{n} - \mathcal{D}^{pre}_{n} \right|,
        \end{equation}
      where $N$ is the number of the testing data, $\mathcal{D}^{label}_{n}$ and $\mathcal{D}^{pre}_{n}$ are the manual read and predicted vessel draft depth, respectively. Theoretically, lower MADDE represents closer result to actual draft depth in our experiments.
      \begin{figure*}[t]
      	\centering
      	\includegraphics[width=1.0\linewidth]{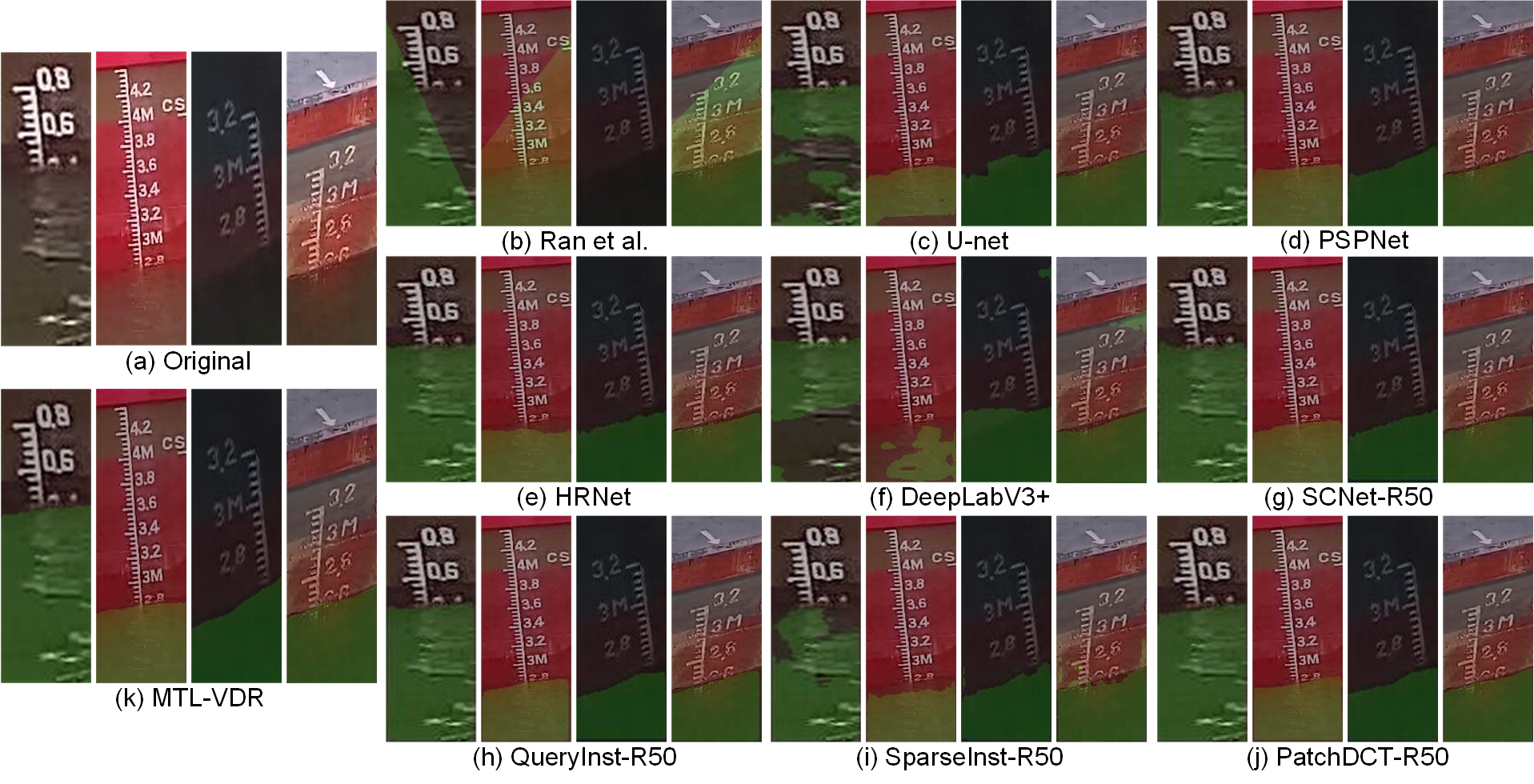}
      	\caption{Visual performance of vessel/water segmentation tasks on clear draft mark images. The green shallow represents the pixels of water. From left-top to right-bottom: The original image (a) and the image segmentation results of (b) Ran \textit{et al} \cite{ran2012draft}, (c) U-net \cite{ronneberger2015u}, (d) PSPNet \cite{zhao2017pyramid}, (e) HRNet \cite{sun2019deep}, (f) DeepLabV3+ \cite{chen2017rethinking}, (g) SCNet \cite{vu2021scnet}, (h) QueryInst \cite{fang2021instances}, (i) SparseInst \cite{cheng2022sparse}, (j) PatchDCT \cite{wen2023patchdct}, and the left-bottom, (k) MTL-VDR. It is noted that the waterline should be a smooth curve or a straight line based on the realistic experience of VDR.}
            \vspace{-0.2cm}
      	\label{Figure_Seg_result_clear}
      \end{figure*}
      \begin{table}[t]
      	\scriptsize
      	\centering
      	\caption{Comparisons with other object detection methods. The image segmentation networks of the compared methods are set to U-net, which is also the baseline of the segmentation subnet in MTL-VDR. The best results are shown in \textbf{bold}. $\downarrow$ represents that lower values denote better results. }
      	\label{Tab:multi-task_det}
      	\begin{tabular}{l|cccc}
      		\hline
      		Method & Type &  Para. (M) & Time (s) & MADDE $\downarrow$ \\ \hline \hline
            \begin{tabular}[c]{@{}l@{}}Faster R-CNN \cite{girshick2015fast}\\+ U-net \cite{ronneberger2015u}\end{tabular}        & Anchor-Based & 52.77         & 0.162          &  1.102 $\pm$ 0.989    \\
      		\begin{tabular}[c]{@{}l@{}}SSD \cite{liu2016ssd} + \\ U-net \cite{ronneberger2015u}\end{tabular}     & Anchor-Based & 34.02         & 0.051        &  1.160 $\pm$ 0.975    \\
      		\begin{tabular}[c]{@{}l@{}}RetinaNet \cite{lin2017focal} + \\ U-net \cite{ronneberger2015u}\end{tabular}     & Anchor-Based & 18.58         & 0.056            &  1.487 $\pm$ 0.965    \\
      		\begin{tabular}[c]{@{}l@{}}YOLOv7-tiny \cite{wang2022YOLOv7}\\+ U-net \cite{ronneberger2015u}\end{tabular}       & Anchor-Based & \textbf{14.98}         & 0.023           &  1.092 $\pm$ 0.898    \\
      		\begin{tabular}[c]{@{}l@{}}FCOS \cite{tian2019fcos} +\\U-net \cite{ronneberger2015u}\end{tabular} & Anchor-Free & 37.26    & 0.062            &     0.460 $\pm$ 0.716\\ 
            \begin{tabular}[c]{@{}l@{}}ATSS \cite{zhang2020bridging} +\\U-net \cite{ronneberger2015u}\end{tabular}        & Anchor-Free & 57.03         & 0.282          &  0.556 $\pm$ 0.784    \\ 
            \begin{tabular}[c]{@{}l@{}}YOLOF \cite{chen2021you} +\\U-net \cite{ronneberger2015u}\end{tabular}        & Anchor-Free & 67.46         & 0.152          &  0.566 $\pm$ 0.789    \\
      		\begin{tabular}[c]{@{}l@{}}YOLOX-s \cite{ge2021YOLOx} +\\U-net \cite{ronneberger2015u}\end{tabular}        & Anchor-Free & 33.83         & 0.024          &  0.541 $\pm$ 0.729    \\ \hline\hline
      		MTL-VDR                     & Anchor-Free     & 19.61         & \textbf{0.016}            &  \textbf{0.074 $\pm$ 0.076}    \\ \hline
      	\end{tabular}
            \vspace{-0.4cm}
      \end{table}
    \subsection{Dataset}
      The proposed MTL-VDR has been applied to a practical maritime surveillance project, which was constructed in Guangdong Province, China. The camera for visual data collection is deployed on a navigation lock in the Shaoguan section of the Pearl River, where over 500 inland vessels pass every month. To verify the effectiveness of our MTL-VDR, we collect and annotate a novel dataset (VDR dataset) for the VDR task. As shown in Fig. \ref{Figure_dataset_collection}, for dataset construction, the raw visual data is first annotated for draft mark detection. Subsequently, image patches of the draft marks are extracted and finely annotated for draft scale recognition and vessel/water segmentation. All of the labels for object detection and image segmentation are kept in VOC format \cite{everingham2010pascal}. Finally, 45 images with various vessels, including 30 clear and 15 stained draft marks, are randomly extracted and labeled for vessel/water segmentation and draft reading accuracy evaluation. The details of our VDR dataset\footnote{ The constructed VDR dataset is available at \url{https://github.com/QuJX/MTL-VDR.}} can be found in Table \ref{Tab:Dataset}. It can be used for training VDR models, which can be deployed on diversified platforms, e.g., navigation lock, bridge pier, unmanned aerial vehicle (UAV), unmanned surface vehicle (USV), etc.
    \subsection{Comparisons with Other Methods}
      We compare our MTL-VDR with some classical and state-of-the-art methods to comprehensively demonstrate its superiority. Specifically, for the comparison of draft scale recognition, we train several object detection networks, including both anchor-based and anchor-free methods, i.e., Faster R-CNN \cite{girshick2015fast}, SSD \cite{liu2016ssd}, RetinaNet \cite{lin2017focal}, YOLOv7-tiny \cite{wang2022YOLOv7}, FCOS \cite{tian2019fcos}, ATSS \cite{zhang2020bridging}, YOLOF \cite{chen2021you}, and YOLOX-s \cite{ge2021YOLOx}. In the matter of vessel/water segmentation, we employ some classical and state-of-the-art image segmentation methods for comparison, i.e., U-net \cite{ronneberger2015u}, HRNet \cite{sun2019deep}, PSPNet \cite{zhao2017pyramid}, DeepLabV3+ \cite{chen2017rethinking}, SCNet \cite{vu2021scnet}, QueryInst \cite{fang2021instances}, SparseInst \cite{cheng2022sparse}, PatchDCT \cite{wen2023patchdct} and the Hough transformation-based method, Ran \textit{et al} \cite{ran2012draft}. The VDR task consists of different subtasks. Therefore, we combine them directly and utilize their results for the evaluation of the efficiency and effectiveness of VDR tasks. It is worth noting that during the comparison experiments of draft scale recognition and vessel/water segmentation, YOLOX-s and U-net are fixed as object detection and image segmentation networks, respectively. These two networks additionally serve as the baseline models for each subnet in our MTL-VDR.
      \begin{figure*}[!thb]
      	\centering
      	\setlength{\abovecaptionskip}{0.1cm}
      	\includegraphics[width=1.0\linewidth]{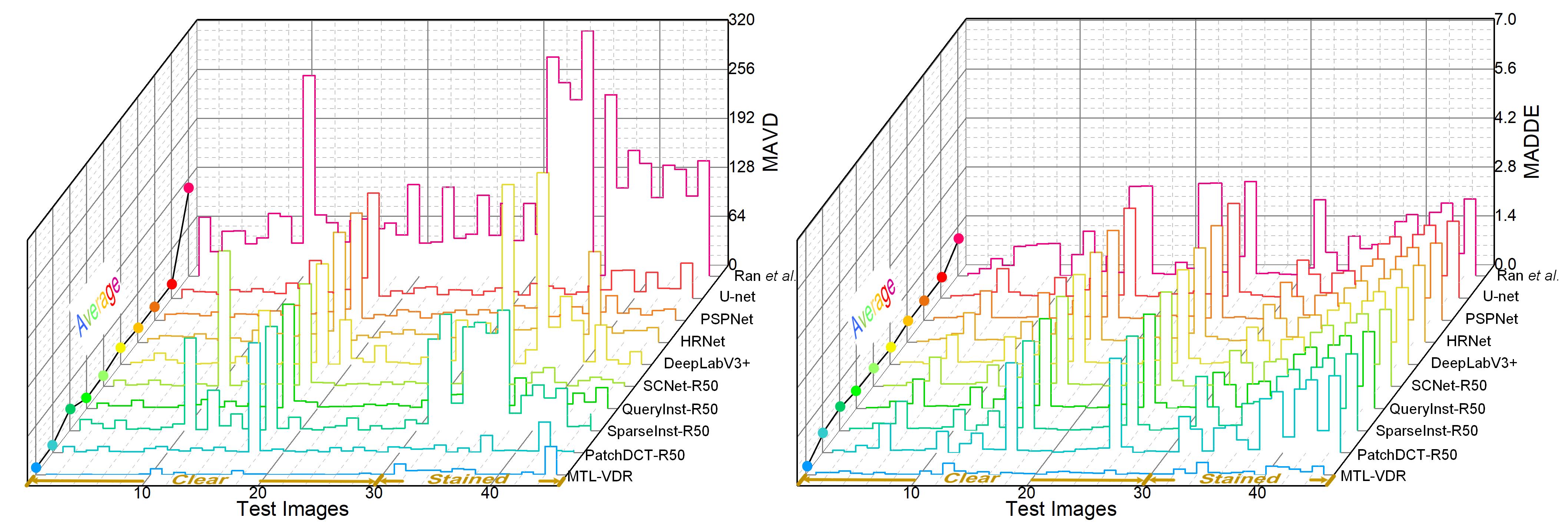}
      	\caption{The quantitative performance of MAVD and MADDE. From inside to outside: the evaluation metric on testing data from Ran \textit{et al} \cite{ran2012draft}, U-net \cite{ronneberger2015u}, PSPNet \cite{zhao2017pyramid}, HRNet \cite{sun2019deep}, DeepLabV3+ \cite{chen2017rethinking}, SCNet \cite{vu2021scnet}, QueryInst \cite{fang2021instances}, SparseInst \cite{cheng2022sparse}, PatchDCT \cite{wen2023patchdct}, and MTL-VDR. The first 30 images have clear draft marks, and the rest 15 are stained. It is noted that the MAVD and MADDE of ground truth (GT) are 0, which represent the same waterline location and reading depth compared with the label, respectively.}
      	\label{Figure_MAVD}
      \end{figure*}
      \begin{table}[t]
      	\scriptsize
      	\centering
      	\caption{Comparisons with other image segmentation methods. The object detection network of the compared methods is set to YOLOX-s, which is also the baseline of the detection subnet in MTL-VDR. The best results are shown in \textbf{bold}. $\downarrow$ represents that lower values denote better results. }
      	\label{Tab:multi-task_seg}
      	\begin{tabular}{l|cccc}
      		\hline
      		Method & Para. (M) & Time (s) & MAVD $\downarrow$ & MADDE $\downarrow$ \\ \hline \hline
      		\begin{tabular}[c]{@{}l@{}}YOLOX-s \cite{ge2021YOLOx} +\\ Ran \textit{et al} \cite{ran2012draft}.\end{tabular}        & \textbf{8.94}       & \textbf{0.016}            & 104.1 $\pm$ 72.83   &  0.887 $\pm$ 0.859     \\
      		\begin{tabular}[c]{@{}l@{}}YOLOX-s \cite{ge2021YOLOx} +\\ U-net \cite{ronneberger2015u}\end{tabular}        & 33.83         & 0.024           & 16.81 $\pm$ 21.03   &  0.541 $\pm$ 0.729   \\
      		\begin{tabular}[c]{@{}l@{}}YOLOX-s \cite{ge2021YOLOx} + \\ PSPNet \cite{zhao2017pyramid},\end{tabular}     & 11.32         & 0.026          & 13.61 $\pm$ 19.78   &  0.538 $\pm$ 0.734    \\
      		\begin{tabular}[c]{@{}l@{}}YOLOX-s \cite{ge2021YOLOx} + \\ HRNet \cite{sun2019deep}\end{tabular}      & 18.58         & 0.063            & 15.93 $\pm$ 25.55   &  0.536 $\pm$ 0.731    \\
      		\begin{tabular}[c]{@{}l@{}}YOLOX-s \cite{ge2021YOLOx} + \\ DeepLabV3+ \cite{chen2017rethinking}\end{tabular} & 14.75         & 0.027            & 28.80 $\pm$ 54.02   &     0.525 $\pm$ 0.742\\
            \begin{tabular}[c]{@{}l@{}}YOLOX-s \cite{ge2021YOLOx} + \\ SCNet-R50 \cite{vu2021scnet}\end{tabular} & 103.41         & 0.197            & 11.97 $\pm$ 31.72   &     0.521 $\pm$ 0.740\\
            \begin{tabular}[c]{@{}l@{}}YOLOX-s \cite{ge2021YOLOx} + \\ QueryInst-R50 \cite{fang2021instances}\end{tabular} & 181.39         & 0.158            & 11.12 $\pm$ 20.28   &     0.522 $\pm$ 0.740\\
            \begin{tabular}[c]{@{}l@{}}YOLOX-s \cite{ge2021YOLOx} + \\ SparseInst-R50 \cite{cheng2022sparse}\end{tabular} & 57.60         & 0.201            & 35.99 $\pm$ 45.40   &     0.556 $\pm$ 0.749\\
            \begin{tabular}[c]{@{}l@{}}YOLOX-s \cite{ge2021YOLOx} + \\ PatchDCT-R50 \cite{wen2023patchdct}\end{tabular} & 134.27         & 0.142            & 8.057 $\pm$ 21.40   &     0.527 $\pm$ 0.737\\
        \hline\hline
        MTL-VDR                         & 19.61         & \textbf{0.016}            & \textbf{3.323 $\pm$ 5.682}   &  \textbf{0.074 $\pm$ 0.076}    \\ \hline
      	\end{tabular}
      \end{table}
      \begin{figure}[t]
      	\centering
      	\setlength{\abovecaptionskip}{0.1cm}
      	\includegraphics[width=1.0\linewidth]{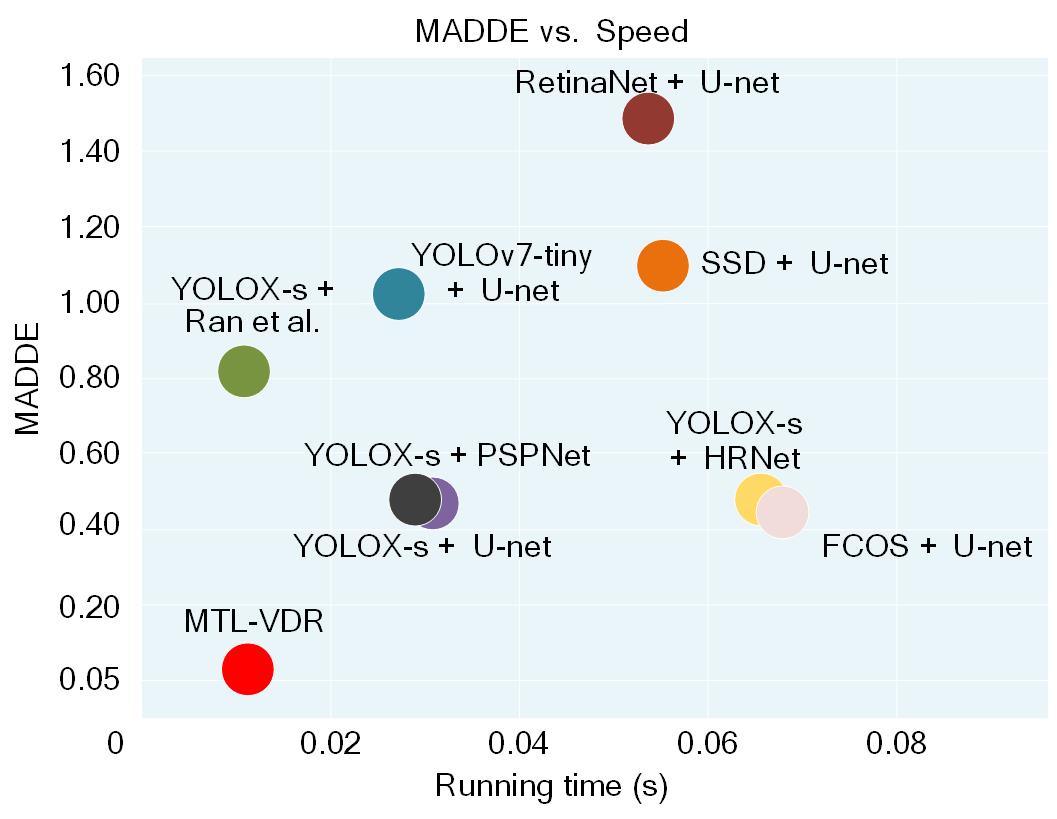}
      	\caption{The trade-off between the MADDE and speed for comparison methods. It is noted that only the methods faster than 10 FPS are depicted. The results show the superiority of our MTL-VDR on both computational efficiency and draft depth estimation accuracy.}
            \vspace{-0.4cm}
      	\label{Figure_MADDE_Speed}
      \end{figure}

      Table \ref{Tab:multi-task_det} presents the experimental comparison about the detection of draft scale characters from various methods. In practical maritime surveillance, the size of the draft scale is diverse due to different imaging conditions, such as varying vessel sizes, different shooting distances, etc. Consequently, it is difficult to generate universal bounding box anchors. As a result, anchor-based object detection approaches tend to perform worse than anchor-free methods, leading to significant deviations from the actual draft depth. Furthermore, our method outperforms the combination of one-stage detectors and U-net \cite{ronneberger2015u} by achieving higher efficiency and greater accuracy in draft depth estimation. This outcome is facilitated by the multi-task learning network architecture, the subsequent draft scale recognition correction, and the accurate vessel/water segmentation.
      The visual performance of the vessel/water segmentation is demonstrated in Fig. \ref{Figure_Seg_result_clear}. Ran \textit{et al}. employed the Canny operator and Hough transformation for waterline detection, but these techniques are susceptible to disruption caused by water waves and color variations in hulls. Additionally, in low-light conditions, the edge feature may not be prominent enough, which further complicates edge detection. U-net, DeepLabV3+, and SparseInst have difficulty in distinguishing between water and vessel hulls with similar colors, leading to erroneous segmentation results. PSPNet, HRNet, and SCNet produce more stable segmentation results on clear draft marks. However, their generated waterlines are unsmooth with many jagged edges. QueryInst and PatchDCT fail to correctly identify water at the edge of the image. The quantitative evaluation results are presented in Table \ref{Tab:multi-task_seg}. Our MTL-VDR performs better on waterline detection, primarily due to the inclusion of $\mathcal{L}_{mavd}$ in the loss function. As a result, our method has obtained more accurate and stable draft reading results when evaluated based on the performance metrics of MAVD and MADDE, as depicted in Fig. \ref{Figure_MAVD}. 
      
      In practical engineering, there are typically multiple tasks that need to be accomplished at the same time. Less computational parameters are beneficial for releasing memory space for other calculations. Moreover, due to the water waves and vessel jolt, the location of the waterline has significant deviations at different times. To tackle this problem, we utilized the average value of the estimated depth in one second as the final draft depth of the vessel. A quicker prediction represents more values to be averaged, and the estimation will be more accurate. Therefore, efficiency is important for VDR methods. Compared with other methods, our MTL-VDR outperforms the combination of YOLOX-s and U-net models by reducing computational complexity with fewer parameters. By analyzing Fig. \ref{Figure_MADDE_Speed}, although some compared methods have smaller model sizes, their use of model switching significantly increases the processing time. In general, our MTL-VDR achieves the best performance on the trade-off between accuracy and efficiency.
      %
    \subsection{Evaluation of Robustness to Stains}
      \begin{figure*}[t]
      	\centering
      	\setlength{\abovecaptionskip}{0.1cm}
      	\includegraphics[width=1.0\linewidth]{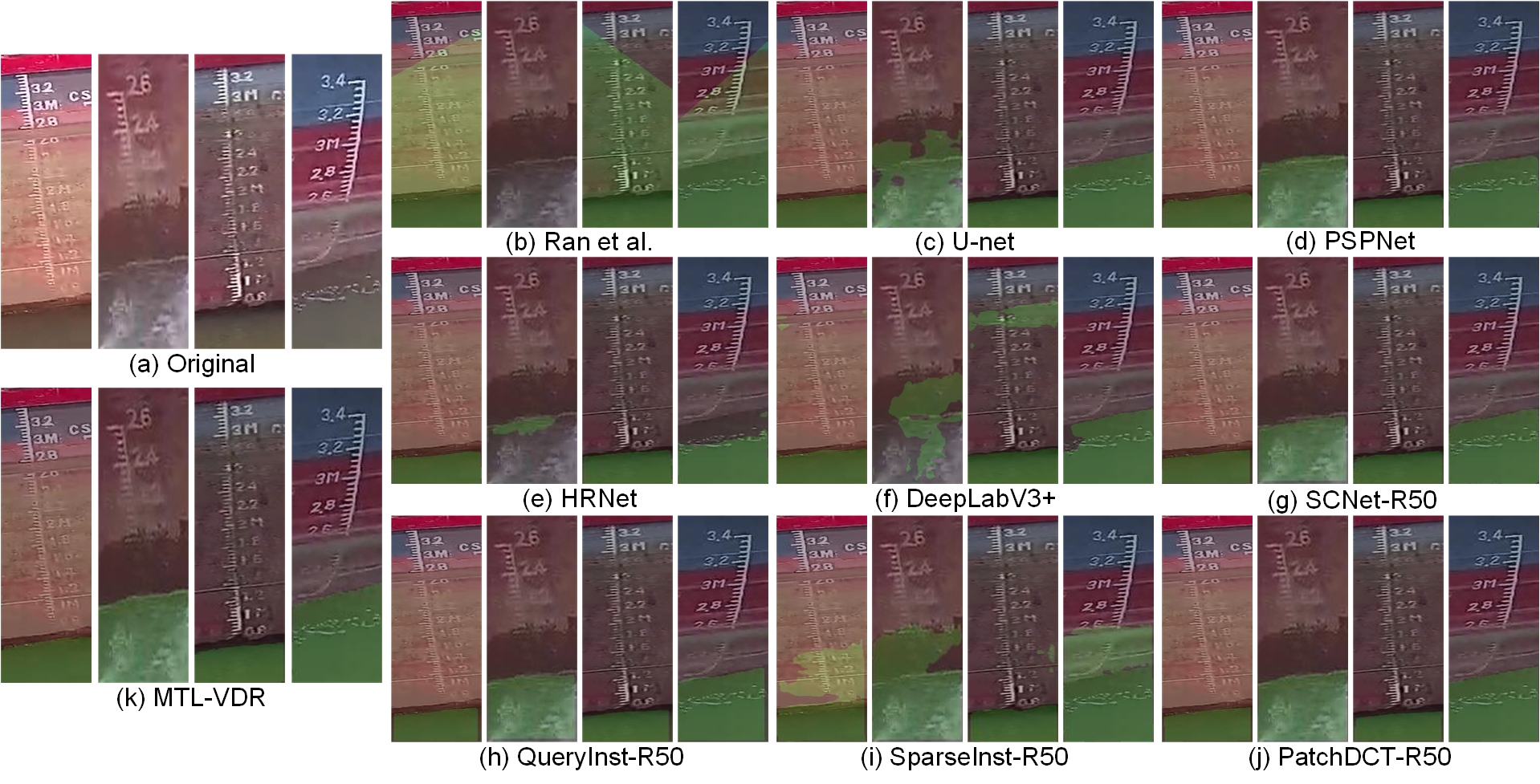}
      	\caption{Visual performance of vessel/water segmentation tasks on stained draft mark images. The green shallow represents the pixels of water. From left-top to right-bottom: The original image (a) and the image segmentation results of (b) Ran \textit{et al} \cite{ran2012draft}, (c) U-net \cite{ronneberger2015u}, (d) PSPNet \cite{zhao2017pyramid}, (e) HRNet \cite{sun2019deep}, (f) DeepLabV3+ \cite{chen2017rethinking}, (g) SCNet \cite{vu2021scnet}, (h) QueryInst \cite{fang2021instances}, (i) SparseInst \cite{cheng2022sparse}, (j) PatchDCT \cite{wen2023patchdct}, and the left-bottom, (k) MTL-VDR. Compared with clear draft marks, the stains on vessel hulls usually have a similar color to the water, which brings a great challenge to the vessel/water segmentation.}
      	\label{Figure_Seg_result_stained}
      \end{figure*}
      \begin{table*}[t]
      	\scriptsize
      	\centering
      	\caption{The ablation study on different components of the proposed method. $\downarrow$ represents that lower values denote better results.}
      	\label{Ablation Study}
      	\begin{tabular}{l|cccc|cc}
      		\hline
      		Method               & $\mathcal{L}_{mavd}$                 & Dynamic Loss Weight       & Scale Correction          & Adaptive Estimation Methods                     & MAVD $\downarrow$ & MADDE $\downarrow$ \\ \hline \hline
      		MTL-VDR      &    \checkmark   &    \checkmark    &   \checkmark    &   \checkmark &  3.32 $\pm$ 5.68  &  0.074 $\pm$ 0.076 \\
      		&    &  \checkmark   &    \checkmark    &   \checkmark   &  9.20 $\pm$ 21.9  &  0.419 $\pm$ 0.651  \\
      		Ablation       & \checkmark &   &   \checkmark  & \checkmark  & 2.70 $\pm$ 5.95 &    0.083 $\pm$ 0.079 \\
      		& \checkmark & \checkmark &   & \checkmark  &   3.32 $\pm$ 5.68  & 0.116 $\pm$ 0.321\\
      		& \checkmark & \checkmark & \checkmark &   &  3.32 $\pm$ 5.68  &  0.108 $\pm$ 0.347 \\ 
      		\hline
      	\end{tabular}
      \end{table*}
      The inland vessel hulls are often covered by sediment and dirt, leading to stains that significantly impact the accuracy of draft scale recognition and vessel/water segmentation. To address this challenge, we propose a draft scale correction method based on their spatial distribution rules. It enables our MTL-VDR to accurately locate and recognize more draft scales on stained draft marks, which provides more reliable results for the final draft depth estimation. We also evaluate the robustness of our proposed MTL-VDR to stained draft marks in terms of vessel/water segmentation experiments. The quantitative experimental results are depicted in Fig. \ref{Figure_MAVD}. Due to the inclusion of MAVD in the loss function, our method performs better in distinguishing vessel hull stains and water with similar colors. It leads to more accurate segmentation results and smoother waterlines, resulting in the estimated vessel draft depth being closer to ground truth with lower MAVD and MADDE. Fig. \ref{Figure_Seg_result_stained} presents the visual comparison results. Ran \textit{et al}. exhibits large deviations in waterline detection due to the complex edge features caused by the covering dirt or water stains. DeepLabV3+ and U-net are sensitive to color variations, resulting in mis-segmentation between stained vessel hulls and water with similar colors. HRNet and SparseInst fail to distinguish water from water marks on the vessel hull. PSPNet, SCNet, QueryInst, and PatchDCT perform better on stained draft marks. However, at the edges and corners of the water object, our MTL-VDR approach produces more accurate segmentation results. Accurate results on both draft scale recognition and vessel/water segmentation benefit robust draft reading with lower MADDE.
    
    \subsection{Ablation Study}
      \begin{table}[t]
      \scriptsize
      \centering
      \caption{Ablation study on the network architecture. In the separate version, our network is trained with a single-task loss function (water/vessel segmentation and draft scale character detection). The trained models are then combined for the final depth estimation.}
      \label{Tab:Ablation_multi_task}
      \begin{tabular}{l|cccc}
      \hline
      Method                    & Para. (M) & Time (s) & MAVD  & MADDE \\ \hline \hline
      \begin{tabular}[c]{@{}l@{}}YOLOX-s \cite{ge2021YOLOx} +\\ U-net \cite{ronneberger2015u}\end{tabular}            & 33.83                         & 0.024                        & 16.81 $\pm$ 21.03                    & 0.541 $\pm$ 0.729                     \\
      \begin{tabular}[c]{@{}l@{}}MTL-VDR\\(Separate Version)\end{tabular} & 26.21                         & 0.034                        &   2.697 $\pm$ 4.378                   & 0.068 $\pm$ 0.072        \\ \hline
      MTL-VDR                   & 19.61                         & 0.016                        & 3.323 $\pm$ 5.682                   & 0.074 $\pm$ 0.076                    \\ \hline
      \end{tabular}
      \end{table}
      To demonstrate the necessity of each component, we design a series of ablation experiments. As shown in Table \ref{Ablation Study}, in the training period of the multi-task learning network, $\mathcal{L}_{mavd}$ enables the segmentation subnet to pay more attention to the pixels of waterlines, which significantly reduces the MAVD. Moreover, the better performance of the segmentation subnet on waterline pixels benefits the accurate draft depth estimation, and the MADDE is thus reduced. The static weight of loss functions results in an imbalance in learning capability for different tasks. Consequently, the MAVD of the static loss function is reduced, but the draft reading result is worse due to bad performance in detecting the draft scale character. The scale correction method is proposed to rectify the incorrect recognition of draft scales. It offers more precise draft scales for ultimate depth estimation. Moreover, for the stained draft marks, it corrects some misdetection results caused by blurred characters. Therefore, our MTL-VDR has better robustness to vessel hull stains with lower MADDE. In most instances, the character height-based draft depth estimation method is feasible. However, the bounding box of the scale is consistently larger than the actual size, and the scales are usually tilted. These problems will cause significant inaccuracies in the final draft depth estimation. The relative distance between different scales is more reliable. Therefore, the dynamically adaptive estimation method improves the reading for various types of draft marks, such as those on lightly or heavily loaded vessels, clear or stained vessel hulls, etc.

      To verify the rationality of the multi-task learning technology in VDR tasks, we also construct experiments for the network architecture. The experimental results are illustrated in Table \ref{Tab:Ablation_multi_task}. Separate training with individual loss functions shows better performance on both MAVD and MADDE. However, the superiority is limited. Meanwhile, due to the more complex computation in separate networks and the time for model switching, the computational time of the separate version is much longer than that of MTL-VDR. It reduces the robustness of the proposed method for dynamic locations of waterlines due to water waves or vessel jolts. Moreover, for the application on moving platforms (UAV, USV, etc.), the efficiency will become more important. For the sake of more accurate and device-friendly performance, MTL-VDR is more appropriate in practical engineering.
\section{Conclusion and future perspectives}
\label{conclusion}
      In this work, we presented a multi-task learning-enabled automatic vessel draft reading method (MTL-VDR). It completes a series of subtasks, i.e., draft mark detection, draft scale recognition, vessel/water segmentation, and final draft depth estimation. To facilitate the detection of image patches with draft marks, we employed the YOLOv8 network. Subsequently, we designed a multi-task learning network that performs draft scale recognition and vessel/water segmentation simultaneously. During training, we took the location of the waterline into consideration and proposed the MAVD loss function to improve the precision of vessel/water segmentation. To increase the robustness of vessel hull stains, we developed a draft scale correction method based on the spatial distribution rules of draft scales. Finally, we employed the dynamically adaptive draft depth estimation method to enhance the accuracy and universality across the applications on different draft marks. Experimental results based on the newly developed VDR dataset demonstrated the effectiveness and efficiency of the proposed method when compared with other competing approaches. Additionally, the results indicated that our approach is robust to vessel hull stains. Ablation studies were constructed on each component of the approach, further demonstrating the reasonable design of MTL-VDR.
      
      In conclusion, our work presents an automatic VDR method for enhancing the efficiency and intelligence of maritime surveillance in ITS. Although our method obtains promising results in this study, it still faces several challenges, e.g., unqualified visual data and limited draft scale recognition accuracy, which will limit the accuracy of VDR results. The further improvement of our method includes the following.
      \begin{itemize}
          \item The inadequate quality of visual data under low-visibility weather, e.g., low lightness and haze, will affect the accuracy of the image analysis algorithm. Therefore, visibility enhancement methods will be considered for better reliability in different weather conditions.
          \item Currently, multi-task learning technology is only employed to reduce computational complexity, which cannot achieve accurate reading results compared with separate networks. In the future, we will try to use it to improve detection and segmentation accuracy simultaneously.
          \item The rule of draft depiction for sea vessels differs from that for inland vessels, and some drafts may be depicted using imperial units. It is thus improper to use the current method to estimate the final draft depth. In further applications, we will optimize the scale recognition and depth estimation approach to broaden the application area in the oceanic industry.
      \end{itemize}
      %
      %
\bibliography{Ref}    
\bibliographystyle{IEEEtran}

\end{document}